\documentclass[sigconf]{aamas}

\usepackage{balance} 

\setcopyright{ifaamas}
\acmConference[AAMAS '26]{Proc.\@ of the 25th International Conference
on Autonomous Agents and Multiagent Systems (AAMAS 2026)}{May 25 -- 29, 2026}
{Paphos, Cyprus}{C.~Amato, L.~Dennis, V.~Mascardi, J.~Thangarajah (eds.)}
\copyrightyear{2026}
\acmYear{2026}
\acmDOI{http://doi.org/10.65109/V1X2Y3Z4}
\acmPrice{}
\acmISBN{}

\acmSubmissionID{<<837>>}

\usepackage{amsmath,amsfonts,bm}

\def\eqref#1{equation~\ref{#1}}

\def\1{\bm{1}}

\DeclareMathAlphabet{\mathsfit}{\encodingdefault}{\sfdefault}{m}{sl}
\SetMathAlphabet{\mathsfit}{bold}{\encodingdefault}{\sfdefault}{bx}{n}

\newcommand{\E}{\mathbb{E}}

\newcommand{\R}{\mathbb{R}}

\newcommand{\Var}{\mathrm{Var}}

\def\cJ{{\mathcal{J}}}

\usepackage{amsmath}
\usepackage{amssymb}
\usepackage{mathtools}
\usepackage{amsthm}
\usepackage{dsfont}

\usepackage{thmtools}

\usepackage{enumitem}

\usepackage{wrapfig}

\usepackage{multirow}
\usepackage{makecell}

\usepackage{microtype}
\usepackage{graphicx}
\usepackage{subfigure}
\usepackage{booktabs} 

\usepackage{hyperref}

\usepackage{amsmath}
\usepackage{amssymb}
\usepackage{mathtools}
\usepackage{amsthm}

\usepackage[capitalize,noabbrev]{cleveref}

\theoremstyle{plain}

\theoremstyle{definition}

\theoremstyle{remark}

\usepackage[textsize=tiny]{todonotes}

\title[AAMAS-2026 Formatting Instructions]{Advancing Multi-Agent RAG Systems \\with Minimalist Reinforcement Learning}

\author{Yihong Wu\textsuperscript{\textdagger}}
\affiliation{
  \institution{Université de Montréal}
  \city{Montréal}
  \state{QC}
  \country{Canada}
  }
\email{yihong.wu@umontreal.ca}

\author{Liheng Ma\textsuperscript{\textdagger}}
\affiliation{
  \institution{McGill University \& Mila}
  \city{Montréal}
  \state{QC}
  \country{Canada}}
\email{liheng.ma@mail.mcgill.ca}

\author{Muzhi Li\textsuperscript{\textdagger}}
\affiliation{
  \institution{The Chinese University of Hong Kong}
  \city{Hong Kong}
  \country{China}
  }
\email{mzli@cse.cuhk.edu.hk}

\author{Jiaming Zhou}
\affiliation{
  \institution{Huawei Noah’s Ark Lab}
  \city{Montréal}
  \state{QC}
  \country{Canada}}

\author{Lei Ding}
\affiliation{
  \institution{University of Manitoba}
  \city{Winnipeg}
  \state{MB}
  \country{Canada}}

\author{Jianye Hao}
\affiliation{
  \institution{Tianjin University}
  \state{Tianjin}
  \country{China}}

\author{Ho-fung Leung}
\affiliation{
  \institution{Independent Researcher}
    \city{Hong Kong}
  \country{China}}

\author{Irwin King}
\affiliation{
  \institution{The Chinese University of Hong Kong}
  \city{Hong Kong}
  \country{China}}

\author{Yingxue Zhang}
\affiliation{
  \institution{Huawei Noah’s Ark Lab}
  \city{Montréal}
  \state{QC}
  \country{Canada}}

\author{Jian-Yun Nie}
\affiliation{
  \institution{Université de Montréal}
  \city{Montréal}
  \state{QC}
  \country{Canada}}

\begin{abstract}
Large Language Models (LLMs) equipped with modern Retrieval-Augmented Generation (RAG) systems often employ multi-turn interaction pipelines to interface with search engines for complex reasoning tasks. 
However, such multi-turn interactions inevitably produce long intermediate contexts, as context length grows exponentially with exploration depth.
This leads to a well-known limitation of LLMs: their difficulty in effectively leveraging information from long contexts.
This problem is further amplified in RAG systems that depend on in-context learning, where few-shot demonstrations must also be included in the prompt, compounding the context-length bottleneck.
To address these challenges, we propose \textbf{Mujica-MyGo}, a unified framework for efficient multi-turn reasoning in RAG.
Inspired by the divide-and-conquer principle, we introduce \textbf{Mujica} (Multi-hop Joint Intelligence for Complex Question Answering), a multi-agent RAG workflow that decomposes multi-turn interactions into cooperative sub-interactions, thereby mitigating long-context issues.
To eliminate the dependency on in-context learning, we further develop \textbf{MyGO} (Minimalist Policy Gradient Optimization), a lightweight and efficient reinforcement learning algorithm that enables effective post-training of LLMs within complex RAG pipelines.
We provide theoretical guarantees for MyGO’s convergence to the optimal policy. Empirical evaluations across diverse question-answering benchmarks—covering both text corpora and knowledge graphs—show that \textbf{Mujica-MyGO} achieves superior performance.

\end{abstract}

\keywords{Reinforcement Learning, Multi-Agent, RAG, QA, LLM}

\newcommand{\BibTeX}{\rm B\kern-.05em{\sc i\kern-.025em b}\kern-.08em\TeX}

\thanks{\textsuperscript{\textdagger}Equally contributed}
\begin{document}


\pagestyle{fancy}
\fancyhead{}


\maketitle 


\section{Introduction}

The advent of Large Language Models (LLMs)~\cite{ouyang2022training,achiam2023gpt} has redefined expectations for Artificial Intelligence, offering new opportunities for many domains.
Despite their versatility, LLMs suffer from hallucination \cite{huang2025survey} and knowledge cutoffs \cite{cheng2024dated}.
While injecting knowledge through continuous learning is a possible solution, it is computationally expensive and risks catastrophic forgetting~\cite{kirkpatrick2017overcoming}.
To address this issue, Retrieval-Augmented Generation (RAG) \cite{guu2020retrieval,lewis2020retrieval}, a technique enabling LLMs to incorporate external knowledge via prompts, has drawn significant attention among researchers.

Most advanced RAG systems~\cite{PoG, Search-o1, RoG, shao2023enhancing, khattab2022demonstrate} employ a multi-turn pipeline to effectively address complex questions. This process typically involves decomposing the initial question into sub-questions, retrieving relevant information for each, and synthesizing a final answer. Such an approach allows for the generation of more fine-grained queries, enabling search engines to return more relevant passages and ultimately enhancing overall performance.

Despite its effectiveness, this multi-turn interaction introduces a significant challenge: the long-context problem. 
For instance, to answer the query, ``\textit{Which U.S.\ presidents previously served as governors?}'', a RAG system must first identify all U.S.\ presidents and then retrieve biographical information to determine which individuals held gubernatorial positions before their presidency. 
Each president has an extensive historical record, and aggregating such information results in a large volume of retrieved text within a single LLM context window. Given that LLMs struggle to capture information accurately in long contexts~\cite{liu2024lost}, this naive aggregation can lead to performance degradation.

Moreover, this issue is compounded by the reliance of In-Context Learning (ICL)~\cite{ICL} to guide the LLM's behavior in most RAG workflows. This technique requires providing few-shot demonstrations within the prompt to instruct the model on the desired task. The performance of ICL is highly dependent on the precision and comprehensiveness of these examples, especially their ability to cover corner cases. Consequently, these detailed demonstrations further inflate the context length, exacerbating the long-context problem.

To address the aggregation problem, we propose \textbf{Mujica} (Multi-hop Joint Intelligence for Complex Question Answering), a multi-agent RAG system designed for complex queries.
Mujica consists of two core modules: a \textbf{Planner} and a \textbf{Worker}.
The Planner initiates the process by analyzing the original query, decomposing it into sub-questions, and delegating them to the Worker.
Based on the responses from the Worker, the Planner iteratively formulates new queries until it has sufficient information to finalize an answer or reaches the iteration limit.
The Worker, in turn, receives sub-questions from the Planner, retrieves relevant passages from search engines, and returns concise summaries. This planner-worker architecture enables a clear separation between high-level reasoning and low-level retrieval.
By maintaining a clean, summary-based history in the Planner's context, Mujica effectively mitigates the long-context problem. 

To eliminate the need for cumbersome few-shot demonstrations in in-context learning (ICL), post-training is a crucial step for a complex RAG system like Mujica.
However, the highly customized nature of such systems means that unified datasets for Supervised Fine-Tuning (SFT) are generally unavailable.
This scarcity of trajectory data naturally frames the post-training process as a Reinforcement Learning (RL) problem, where the RAG system constitutes the policy and its performance on a task determines the reward.
Nevertheless, optimizing this policy via RL is non-trivial due to the intricate interactions between the agents and the search engine.
To tackle this challenge, we introduce Minimalist policy Gradient Optimization (\textbf{MyGO}), a simple and efficient RL algorithm specifically designed for such complex RAG systems.

\begin{figure*}[t]
	\centering
	\includegraphics[width=\textwidth]{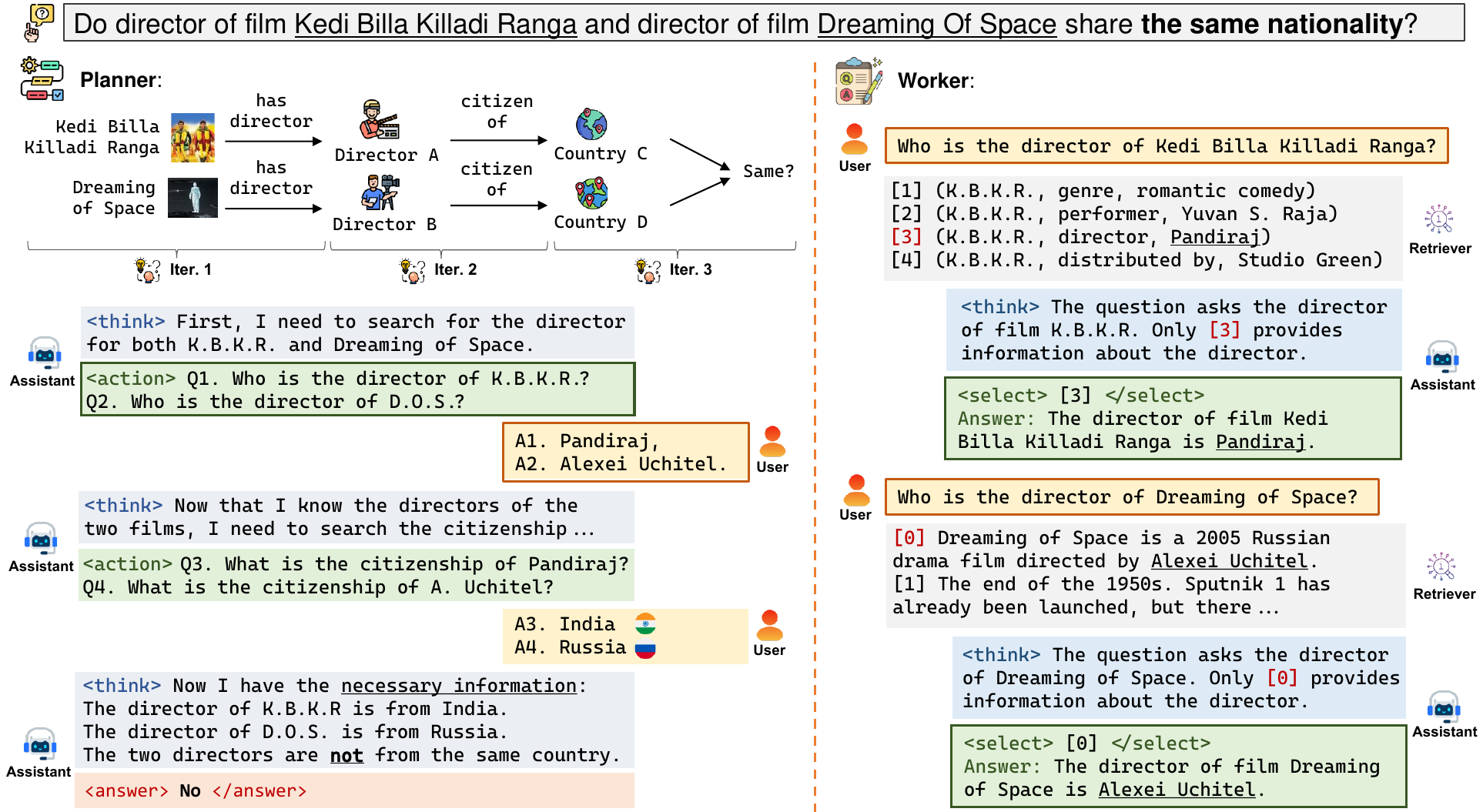}
	\caption{The end-to-end architecture of the proposed Mujica framework.}
	\label{pipeline}
\end{figure*}

We designed MyGO to be minimalist, deliberately avoiding components like reward weighting, importance sampling, or token clipping that are common in other algorithms.
This approach stands in contrast to methods like Proximal Policy Optimization (PPO)~\cite{schulman2017proximal}, which requires an auxiliary value network to stabilize training, and Group Relative Policy Optimization (GRPO)~\cite{shao2024deepseekmath}, which must generate multiple trajectories per query for reward normalization, exacerbating latency.
The key innovation in MyGO lies in its sampling strategy. By sampling trajectories from an asymptotically approximate optimal policy, we can directly use Maximum Likelihood Estimation (MLE) for policy updates.
This insight allows MyGO to be implemented easily within standard SFT frameworks, lending it high stability during training and simplifying hyperparameter tuning.
We provide a detailed theoretical justification for the validity of this approach and show through empirical experiments on various datasets that Mujica-MyGO effectively improves RAG performance across different LLMs.

\section{RAG Agent}


\subsection{Problem Definition}

This work addresses Open-Domain Multi-Hop Question Answering (MHQA), a key challenge and benchmark for RAG.
QA systems aim to provide correct answers to given questions.
Compared with single-hop QA, MHQA necessitates reasoning across multiple sources (hops)~\cite{mavi2024multi}, which renders MHQA a more challenging task. 
We focus on the open-domain setting, where relevant information must be retrieved from large-scale corpora (knowledge graphs or text collections).
Details are provided in Appendix~\ref{probelm_definition}.

The inherent complexity of MHQA, characterized by multi-stage planning requirements and iterative evidence synthesis, necessitates framing the solution mechanism as an autonomous agent.
We define an agent as an entity capable of planning, taking actions based on the environment, and engaging in sequential reasoning \cite{liu2025advances}.
We observe that LLMs equipped with RAG perform functions analogous to such an agent. Specifically, LLMs determine information requirements and search strategies (planning), formulate search queries (acting), and synthesize final answers from gathered evidence (reasoning). Recognizing this functional correspondence, we position the LLM itself as the core agent within our system.

\subsection{Mujica}
As previously noted, repeated retrieval from RAG systems generates extended contexts that are challenging for LLMs to effectively process.
To mitigate this, a straightforward approach is to adopt a divide-and-conquer approach by partitioning the context into smaller segments.
Accordingly, we introduce \underline{\textbf{M}}ulti-hop \underline{\textbf{J}}oint \underline{\textbf{I}}ntelligence for \underline{\textbf{C}}omplex Question \underline{\textbf{A}}nswering (\textbf{Mujica}) -- a multi-agent RAG framework that decomposes multi-turn interactions into coordinated sub-interactions.

Figure~\ref{pipeline} shows the overall pipeline of the proposed agentic RAG framework. Given a complex question $q$, our RAG agent aims to decompose it into a series of simple subquestions $\{q_1, q_2, \cdots, q_n\}$, and progressively answers each subquestion based on their internal dependencies until obtaining the final answer $a$. Let $\mathcal{I}$ denote the instruction prompts, $\mathcal{C} = \{
\mathcal{C}_1, \mathcal{C}_2, \cdots, \mathcal{C}_n\}$ denote the supporting contexts of each subquestion $q_i$, and $a_i$ an answer to it. Inspired by~\cite{Search-o1}, we can formulate the answer generation process as the following objective function: 
\begin{equation}~\label{decoupling}
\begin{aligned}
   &f_{\text{QA}}(q,I,\mathcal{C}) = P(a|q, I,\mathcal{C}) \\ &= \prod_{i=1}^n{P(q_i|a_{<i}, q_{<i}, q, \mathcal{I})} \cdot \prod_{i=1}^n{P(a_i|q_i,\mathcal{C}_i,\mathcal{I})},
\end{aligned}
\end{equation}
where $q_{<i}$ and $a_{<i}$ denote all preceding subquestions of subquestion $q_i$ and their corresponding answers. Based on the equation above, 
the QA framework can naturally be decoupled into two specialized roles:
\begin{enumerate}[label=\arabic{enumi})]
    \item \emph{Planner}: {Being responsible for planning how to answer a complex question, the planner determines the series of subquestions to be addressed, identifies what specific information needs to be retrieved, and adjusts the plan based on the answer of each subquestion. } 
    \item \emph{Worker}: {Acting as a mini-RAG system, the worker directly interacts with the retriever and tackles specific sub-questions identified by the planner. }
\end{enumerate}

\emph{Planner} and \emph{worker} interact with each other through a conversational process, each serving as both user and assistant for one another. 
The planner and worker may operate as independent agents (distinct LLMs) or as virtual agents backed by a shared LLM but conditioned on distinct instruction prompts. 
Considering the substantial size of LLMs, we opt for the latter design for better computational efficiency.

\paragraph{Planner: Planning Subquestions as a Directed Acyclic Graph.}



\begin{wrapfigure}{r}{0.4\linewidth}
	\vspace{-1em}
	\centering
	\includegraphics[width=1\linewidth]{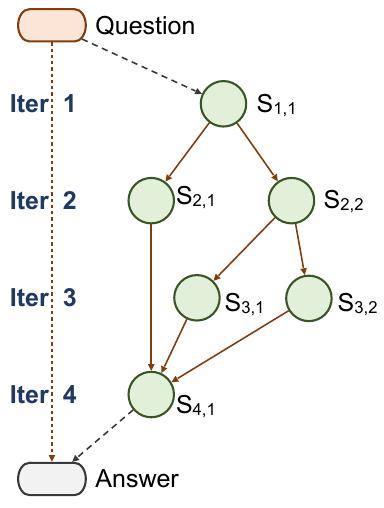}
	\caption{Modeling Complex Question Answering Process as a Directed Acyclic Graph.}
	\label{fig:qdag}
    \vspace{-1em}
\end{wrapfigure}

Recent work (e.g., Search-o1~\cite{Search-o1}, PoG~\cite{PoG}) has demonstrated that performing chain-of-thought reasoning through Eq.~\ref{decoupling} is capable of deriving correct answers.
However, in real-world scenarios, the subquestions posed are not naturally shaped as a sequence, since not all subquestions depend on answers to preceding ones.
As shown in Fig.~\ref{fig:qdag}, there might be two conditionally independent subquestions, $\text{S}_{2,1}$ and $\text{S}_{2,2}$, which are dependent in the later subquestion $\text{S}_{4,1}$.
The dependency relations form a directed acyclic graph (DAG).
Directly applying Eq.~\ref{decoupling} to such a reasoning process can be both inefficient and suboptimal.
Therefore, to effectively handle DAG dependency graphs,
we allow our Mujica planner to ask subquestions in multiple iterations sequentially, where the answers of subquestions at an iteration can become dependent in later iterations.
In each iteration, Mujica will simultaneously ask multiple conditionally independent subquestions.

More specifically, following the previous work~\cite{ReAct},
in each iteration,  the planner will execute two steps:
\textbf{\emph{think}} and \textbf{\emph{action}}.
In the \textbf{\textit{think}} step, at the $i$-th iteration, the planner summarizes the information obtained in the previous iterations and judges whether the supporting information it has gathered is sufficient to answer the complex question $q$. 
If all necessary information is available, the planner generates its conclusion for the entire complex question.
Otherwise, the planner will excute the \textbf{\emph{action}} step - it formulates the subquestions and outsources them to a \textbf{\emph{Worker}} to retrieve the corresponding information from the environment. 
After gathering the information from the \textbf{\emph{Worker}}, the \textbf{\emph{Planner}} will move to the next iteration, and this process continues until obtaining the final answer.

\paragraph{Worker: Interacting with the External Environments.}

As aforementioned, the \textbf{\emph{Worker}} is responsible for handling interactions with the external environments, and answers questions assigned by the planner. 
Specifically, we utilize the same LLM-agent with a mini-RAG system.
For each simple question $q_i$, the worker invokes an external retriever to fetch the top $k$ most relevant supporting contexts from a KG or a document corpus.~\footnote{We employ bge-large-en-v1.5 sentence transformer as the external retriever. 
}
Then, the \textbf{\emph{Worker}} employs the LLM to review the target of the question, examine each retrieved context and accordingly anwser the subquestions outsourced by the \textbf{\emph{Planner}}. Specifically, the LLM is instructed to explicitly illustrate its thoughts, and select  relevant contexts by providing their indexes.
Detailed prompts for \textbf{\emph{Planner}} and \textbf{\emph{Worker}} are provided in Appendix~\ref{appendix_working_examples}.



\section{RL for RAG Agent}



\begin{figure*}
    \centering
    \includegraphics[width=\textwidth]{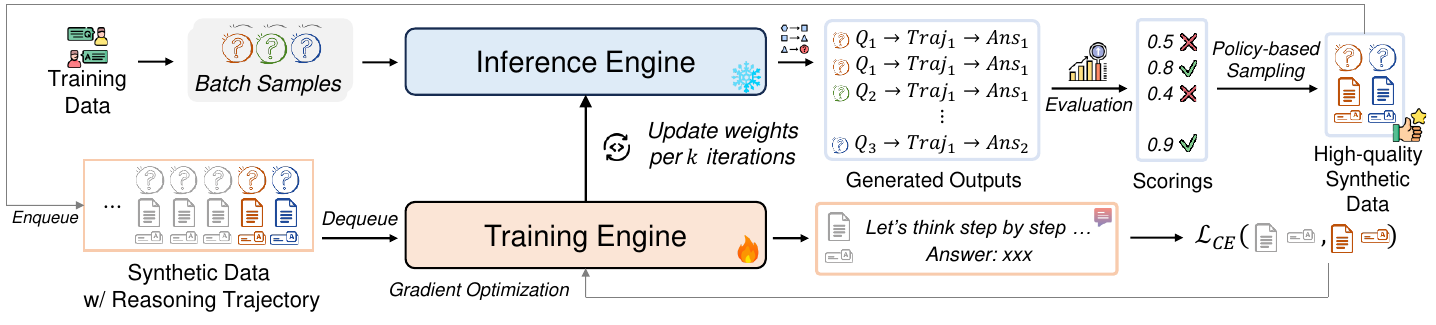}
    \caption{The Proposed Minimalist Policy Gradient Optimization Framework.}
    \label{Mygo}
\end{figure*}

\subsection{Reinforcement Learning Setting}
Our RAG agent is implemented within a conversational framework.
A conversation is modeled as a finite sequence of interactions between the environment and the agent, denoted as $(e_0, a_0, e_1, a_1, \dots, e_T, a_T)$.
The initial environment message $e_0$ comprises the user's question and any necessary system prompts.
For each turn $t \in [0, T]$, $a_t$ is the agent's action, and $e_{t+1}$ is the subsequent environment message.
We define the state $s_t$ at turn $t$ as the history of all preceding interactions, $s_t = (e_0, a_0, e_1, a_1, \dots, a_{t-1}, e_t)$.
Then, a trajectory $\tau = (s_0, a_0, s_1, a_1, \dots, s_T, a_T)$ to the final answer of the question can be denoted as the sequence of states and actions in each step. 
This formulation adheres to a Markov Decision Process (MDP).
Since the golden reasoning path to the ground truth answer is not available in real-world scenarios and most datasets,
 we do not have supervision signals for each action $a_t$ to conduct supervised finetuning.
The only supervision signal available to estimate the reward $r(\tau) \in \mathbb{R}$ for the entire trajectory $\tau$ is the final answer, which naturally corresponds to the terminal reward in an RL setting.



We consider a policy-based method - the RAG agent adopts a policy $\pi_\theta$ to generate reasoning trajectories, where the parameters $\theta$ are optimized using RL techniques.
The objective of the RL is to find  the best policy that maximizes the expected cumulative reward:
\begin{equation}
\label{eq_obj}
     \mathcal{J}(\theta) = \mathbb{E}_{\tau \sim \pi_{\theta}} [r(\tau)] \; ,
\end{equation}
If trajectories sampled from $\pi_\theta$ are differentiable w.r.t. $\theta$, the gradient of the optimization objective can be estimated using the Vanilla Policy Gradient (VPO)~\citep{williams1992simple}:
\begin{equation}
\label{eq_vpo} 
\nabla_\theta \mathcal{J}(\theta) = \underset{\tau \sim \pi_{\theta}}{\mathbb{E}}
\left[ r(\tau) \sum_{t=0}^T \nabla_\theta \log \pi_\theta (a_t | s_t) \right] \;.
\end{equation}


\subsection{Reinforcement Learning for LLMs}


Even though the VPO (Eq.~\ref{eq_vpo}) estimator is unbiased,
it often exhibits high gradient variance, which potentially renders the training process unstable.
To address this issue, many previous approaches \cite{schulman2015trust, schulman2017proximal} propose to replace the raw trajectory reward $r(\tau)$ with the Generalized Advantage Estimation (GAE) \citep{schulman2016high}.
These RL techniques, particularly Proximal Policy Optimization (PPO)~\cite{schulman2017proximal} and its variants, have been successfully applied in the post-training process for LLM~\citep{ouyang2022training, rafailov2023direct}.
Recently, Group Relative Policy Optimization~(GRPO) \citep{shao2024deepseekmath} has been introduced as a simplified alternative to PPO, reducing the training costs while maintaining comparable performance.
The objective function of PPO and GRPO, as adapted for our QA context, can be written as follows:
\begin{equation}
\label{eq_ppo}
\begin{aligned}
    \mathcal{J}(\theta) = &\underset{\tau \sim \pi_{\theta_k}}{\mathbb{E}}
    \left[ A(\tau) \sum_{t=0}^T
\min \left(\mathbf{r}_t, \;
\text{clip}\left(\mathbf{r}_t, 1-\epsilon, 1+\epsilon \right) \right)\right]
\\& \hspace*{\fill} - \beta \mathds{D}_{\text{KL}}(\pi_\theta || \pi_{\text{ref}})  \; ,
\end{aligned}
\end{equation}
where $\epsilon, \beta$ are hyperparameters; $\pi_{\text{ref}}, \pi_{\theta_k}$ denote the reference policy and the behavior policy, respectively; 
$\mathbf{r}_t = \frac{\pi_\theta(a_t|s_t)}{\pi_{\theta_k}(a_t|s_t)}$ is the importance sampling rescaling factor; and $A(\tau)$ is the advantage of applying trajectory $\tau$.
Specifically, in PPO, the advantage function $A(\tau)$ is estimated by $r(\tau) - V(\tau)$, relying on a learnable value function $V(\tau)$.
In GRPO, $A(\tau)$ is instead estimated by $\frac{r(\tau) - \Bar{r}}{\text{std}(r)}$, where $\Bar{r}$ and $\text{std}(r)$ are the mean and standard deviation of $r(\tau)$ for each $\tau$ obtained from the same sampling group, respectively. 

\subsection{MyGO}
\label{sec:mygo}

Despite its wide adoption, a PPO-style objective has two shortcomings: hyperparameter sensitivity and training inefficiency.
The $\epsilon$ hyperparameter is highly sensitive - a small $\epsilon$ may ignore most available trajectories during training, leading to training inefficiency; while a large $\epsilon$ might lose control over the importance sampling scale, resulting in a less stable training process.
The advantage estimation renormalizes the original rewards relative to either the value functions or the group average, 
which weakens the supervision signals from trajectories with large rewards that are helpful for training~\cite{liu2024decade}.
It is worth mentioning that \emph{training inefficiency and instability are significant challenges when training LLMs, which is both computationally expensive and time-consuming.}
On the other hand, compared with previous RL scenarios in robot control and video games~\cite{openai2016gym}, the LLM setting poses a new characteristic: fast simulation, deterministic environment, and trajectory-level reward~\cite{li2024remax}.
In this scenario, \emph{sampling desired trajectories from the environment is relatively more efficient.} 


To this end, we propose our new RL algorithm, \textbf{MyGO}: \textbf{M}inimalist Polic\textbf{y} \textbf{G}radient \textbf{O}ptimization, which shifts the burden from model training to trajectory sampling,
in which we can fully leverage the fast and relatively inexpensive simulation of LLMs~\cite{kwon2023efficient, zheng2024sglang}. 
Unlike previous methods, MyGO disentangles the RL training process into two phases: sampling and training.
During sampling, MyGO directly samples trajectories from the asymptotically optimal distribution and consequently allows the use of Maximum Likelihood Estimation (MLE) to optimize the policy function.
Notably, optimizing a model via MLE is simple, stable, and well established as a regular training technique widely used in unsupervised pre-training~\cite{radford2018improving} and supervised fine-tuning~\cite{ouyang2022training}.


\paragraph{Sampling}
\renewcommand{\cJ}{\mathcal{J}}
We first characterize the target optimal distribution $\pi^*$.
Alternative to the original objective $\mathcal{J}$, we consider the expected reward with entropy regularization $\mathcal{J}' = \E[r(\tau)] + \alpha \mathbf{H}(\pi)$ with $\alpha \in \mathbb{R}^+$,
which, as a Boltzmann distribution \cite{ziebart2010modeling}, has a closed-form solution for the corresponding optimal policy, denoted as ${\pi'}^*$:
\begin{equation}
    \label{eq_boltzmann_optimal} 
    \begin{aligned}
    {\pi'}^* (\tau) &= \frac{\exp (r(\tau)/\alpha)}{Z(\alpha)} \; ,  \\ \quad \text{where } Z(\alpha) &= \int_{\mathcal{T}} \exp (r(\tau')/\alpha) d\tau' \; ,
    \end{aligned}
\end{equation}
where $\alpha$ is the temperature and $Z(\alpha)$ is the partition function that normalizes the distribution over the space of all possible trajectories $\mathcal{T}$.
With a sufficiently small $\alpha \to 0$, we have $\cJ' \to \cJ$, thus ${\pi'}^* \to \pi^*$.
As $\alpha \to 0$,
${\pi'}^*$ becomes sharp and highly skewed,
concentrating its probability mass on trajectories yielding high rewards.
Therefore, given a sufficiently large threshold $K$ close to $\sup_{\tau \in \mathcal{T}} r(\tau)$,
sampling trajectories $\tau$ with $r(\tau) > K$ is asymptotically equivalent to sampling from the ${\pi'}^*$ and thus effectively approximating to sampling from the optimal policy ${\pi}^*$ of $\cJ$.

In practice, we relax the choice of $K$ by setting an update role of $K:= \max(K', (\bar{r}_n/(r_\text{sup} + 1))\cdot r_\text{sup})$, where $r_\text{sup}$ denotes $\sup_{\tau \in \mathcal{T}} r(\tau)$, $\bar{r}_n$ is the empirical average reward of a batch of $n$ sampled trajectories, and $K'$ is the previous threshold value which is initialized as a moderate value depending on the choice of dataset.
As the policy $\pi_\theta$ improved,  $K$ will be asymptotically close to $\sup_{\tau \in \mathcal{T}} r(\tau)$, thus making the sampling distribution close to ${\pi'}^*$.

We highlight that this approach is practically feasible for the reward landscapes for tasks of the LLM agent, which differ significantly from those in complex robotic control problems.
For LLM agents, trajectories frequently culminate in outcomes that are clearly classifiable as successful or unsuccessful.
Empirically, we observe that well-prompted LLMs can readily generate a significant proportion of such successful trajectories, which makes directly sampling from $\pi^*$ feasible.






Let $\pi^{>K}$ denote the distribution that consists of trajectories with $r > K$.
To clarify, we provide two propositions as follows, where the proofs can be found in Appendix~\ref{appendix_math_proof}:
\begin{restatable}{proposition}{klapprox}
\label{propos_1}
Given $\delta \in \R$, $\alpha \in \R$, there exists $K \in \R^+$ such that the following inequality holds:
\begin{equation*}
    \mathds{D}_{\text{KL}}(\pi^{>K} || \pi^*) < \delta \; 
\end{equation*}
\end{restatable}

\begin{restatable}{proposition}{varapprox}
\label{propos_2}
The following property holds when $\alpha$ approaches $0$:
    \begin{equation*}
        \underset{\pi^{>K}}{\Var[r]} \propto 1 / Z^{>K}(\alpha)^2 \; , \; Z^{>K}(\alpha) = \int^{\sup(r)}_K \exp(r/\alpha)dr
    \end{equation*}
\end{restatable}

Proposition~\ref{propos_1} states that if $K$ is sufficiently large, then $\pi^{>K}$ can always approximate $\pi^*$.
Proposition~\ref{propos_2} says the variance of $\pi^{>K}$ is small, considering that $Z$ is usually a large value.
These two propositions guarantee the proximity and stability of our proposed sampling strategy.



\paragraph{Learning}
Once a sufficiently large dataset of optimal trajectories $\mathcal{D}$ has been collected, the optimization of policy $\pi_\theta$ can be simplified to maximizing the log-likelihood $\mathcal{L}$ of these trajectories.
The learning process can be conducted in an online or an offline manner.
This yields the following MLE gradient:
\begin{align}
\label{eq_8}
    \nabla_\theta \mathcal{L} &=\frac{1}{|\mathcal{D}|}\sum_{\tau_i \in \mathcal{D}} \nabla_\theta \log \pi_\theta (\tau_i) \notag \\
    &= \frac{1}{|\mathcal{D}|}\sum_{\tau_i \in \mathcal{D}}
      \sum_{j=0}^{|\tau_i|} \nabla_\theta \mathds{1}(o_j^i) \log \pi_\theta (o_j^i | o_{<j}^i) \; ,
\end{align}
where $o_j^i$ denotes the $j$-th token from $\tau_i$ and $o_{<j}^i$ denotes the sequence of tokens preceding $o_j$, $\mathds{1}(o_j)$ is a characteristic function indicating whether $o_j$ is a token from the agent.
The last equality holds because the action likelihood in the language setting could be further elaborated with token likelihood, i.e, $\log \pi_\theta (a_t | s_t) = \sum_j \log \pi_\theta (o_j | o_{<j})$.

This MLE objective, Eq.~\ref{eq_8}, is equivalent to a standard cross-entropy loss function, widely used in both unsupervised pre-training \cite{radford2018improving} and supervised fine-tuning \cite{ouyang2022training} of language models.
Compared to PPO-style algorithms, Eq.~\ref{eq_ppo}, MLE does not incorporate an advantage function, policy ratio clipping, and importance sampling.
Similarly, unlike methods such as VPO, Eq.~\ref{eq_vpo}, MLE directly uses the curated data $\mathcal{D}$ for training and does not explicitly use reward values in the loss computation.
We empirically observed that MLE leads to more stable training. We explain this increased stability by several factors: (i) the dataset $\mathcal{D}$ consists of trajectories sampled from a more stationary (approximated optimal) distribution, reducing variance compared to on-policy RL where the data distribution shifts rapidly; (ii) the direct use of a log-likelihood objective (cross-entropy) is often well-behaved and less sensitive to reward scaling issues that can affect RL algorithms; and (iii) the avoidance of importance sampling ratios, which can introduce instability if they become too large or small.
The implementation details are in Appendix~\ref{appendix_implementation_detail}.

Compared with previous works, like RAFT~\cite{dongraft} and STaR \cite{zelikman2022star}, which try to simplify RL in LLM post-training, we use a progressive threshold to select data.
More importantly, we provide a theoretical justification that elucidates why this approach, yet relatively simple, yields strong performance, offering novel insights into its efficacy.








\section{Experiments}


\subsection{Environment Design}
\paragraph{Datasets and Evaluation Metrics}
To evaluate the effectiveness of our proposed agent workflow (Mujica) and training method (MyGO) on complex question answering tasks, we utilize four representative benchmark datasets: 2Wiki-MultihopQA (2Wiki) \citep{ho2020constructing}, QALD-10 \citep{QALD-10}, HotpotQA \citep{yang2018hotpotqa}, and MuSiQue \citep{trivedi2021musique}.
2Wiki and QALD-10 are KBQA benchmarks; HotpotQA and MuSiQue are text-based QA dataset.
We use 2Wiki and HotpotQA for training, QALD-10 and MuSiQue for zero-shot evaluation.
We assess QA performance using Exact Match (EM) and F1 score.
For each dataset, these metrics are computed using the official evaluation scripts provided by the respective dataset creators.
Comprehensive dataset statistics and definitions of the evaluation metrics are detailed in Appendix~\ref{appendix_datasets}.
Our experiment is conducted in an online manner.
Due to the limited space, we leave the discussion of offline experiment in Appendix~\ref{appendix_offline}.

\vspace{-1em}
\paragraph{Environment Setting}
To evaluate our agent's capabilities across different data modalities, we derive two experimental environments from the 2Wik dataset: 2Wiki-KG and 2Wiki-Text.
This setup allows us to evaluate the agent's performance on both structured (KGs) and unstructured (text) data.
We use the F1 score as the reward function for all environments.

The \textbf{2Wiki-KG} environment is centered around the Wikidata KG \citep{Wikidata}.
A knowledge graph (KG) is composed of factual statements represented as triples, such as (\textit{Donald Trump}, \textit{position held}, \textit{President of the United States}).
When the agent visits an entity, it can access all triples for which that entity is the head.
The agent's task involves traversing the KG by interpreting the semantic relationships encoded in these triples.
For example, to answer the question "\textit{Who is the mother of the director of film Polish-Russian War (Film)?}", the agent must identify and follow the correct reasoning path, such as: \textit{Polish-Russian War, director, Xawery Żuławski, mother, Małgorzata Braunek}.
To emphasize the reasoning aspect, we follow prior work \citep{ToG, ToG2} by providing the agent with the gold topic entities extracted from the dataset as the entry.


We establish the \textbf{2Wiki-Text} environment to further evaluate our agent's ability to interact with text-based search engines.
The original 2Wiki dataset provides ten supporting passages for each question, which contain the necessary information to derive the answer and the "distractor setting" is designed to test multi-hop reasoning abilities without retrieval.
In our 2Wiki-Text environment, we adapt this by aggregating all passages to form a large corpus.
This approach, following  prior work \citep{IRCoT, aguerrebere2023similarity}, simulates an open-domain retrieval task.
The agent's objective is to answer questions by issuing search queries  and then synthesizing information from the retrieved passages.


Beyond the 2Wiki environments, we extend our text-based evaluation using the HotpotQA dataset, from which we construct two additional settings: \textbf{Hotpot} and \textbf{Hotpot-Kimi}.
In \textbf{Hotpot}, we follow the vanilla setting that uses the Wikipedia dump provided by the dataset~\cite{yang2018hotpotqa}, which only includes the title and the introductory paragraph.
However, we found that the  retriever~\cite{karpukhin2020dense} is the bottleneck for QA performance in the HotpotQA dataset.
To isolate and more directly assess the agent's reasoning capabilities independent of retrieval imperfections, we introduce the \textbf{Hotpot-Kimi} environment.
Similar to but different from the distractor settings~\citep{yang2018hotpotqa, ho2020constructing}, in the {Hotpot-Kimi} environment,the agent is directly provided with the ten gold supporting passages associated with each question, thereby guaranteeing access to all necessary evidence.
However, its primary purpose here is to establish an idealized retrieval condition.
Therefore, the gold passages are not directly fed into the QA-agent like distractor setting. 
Detailed examples and prompts for each environment are provided in Appendix~\ref{appendix_working_examples}.

\subsection{Main Results}
\subsubsection{\textbf{Effectieness of MyGO}}
We evaluate our RAG agent under three distinct settings: Few-Shot, WarmUp, and MyGO. In the \textbf{Few-Shot} setting, the agent is provided with a small number of illustrative examples via in-context learning within the prompt.
Following common practice, we then perform a \textbf{WarmUp} phase.
This phase is to adapt the base model beyond reliance on few-shot exemplars.
For this purpose, we fine-tune the model on 1k samples with EM of 1.
The \textbf{MyGO} training then commences, using this warmed-up model as its initialization.
We conduct our main experiments with the \texttt{Qwen2.5-7B-Instruct} (Qwen) LLM. The feasibility of using \texttt{Llama-3.1-8B-Instruct} (Llama) as the backbone LLM will also be discussed.

\begin{table}[h]
\centering 
\renewcommand{\arraystretch}{1.1} 
\caption{Comparative Performance of Models on 2Wiki-KG and 2Wiki-Text Environments.}
\label{tab:2wiki} 
\resizebox{0.45\textwidth}{!}{
\begin{tabular}{@{}llcccc@{}} 
\toprule
\textbf{Type} & \textbf{Model} & \multicolumn{2}{c}{\textbf{2Wiki-KG}} & \multicolumn{2}{c}{\textbf{2Wiki-Text}} \\
\cmidrule(lr){3-4} \cmidrule(lr){5-6} 
 &  & \textbf{EM} & \textbf{F1} & \textbf{EM} & \textbf{F1} \\
\midrule
\multirow{1}{*}{Few-Shot} & GPT-4.1 & 78.50 & 84.24 & 30.70  & 52.14 \\
\midrule
Few-Shot & \multirow{3}{*}{Qwen2.5-7B}& 57.60 & 62.24 & 23.35 & 33.52 \\
WarmUp &  & 74.93 & 80.74 & 50.18 & 56.25 \\
MyGO&  & 77.63 & 84.15 & 53.17 & 59.62 \\
\midrule
Few-Shot & \multirow{3}{*}{Llama3.1-8B}& 77.70 & 82.09 &  37.83&  46.53\\
WarmUp &  &  82.00&  85.90& 58.03 & 64.81 \\
MyGO&  &  85.93&  91.61&  58.88&  65.88\\
\bottomrule
\end{tabular}
}
\end{table}

Table~\ref{tab:2wiki} presents the performance of these models on the 2Wiki dataset.
Agents trained with MyGO consistently demonstrate strong performance.
In the 2Wiki-KG environment, MyGO-Qwen achieves performance on par with a few-shot prompted GPT-4.1, while the Llama model significantly surpasses this baseline.
In the 2Wiki-Text environment, both MyGO-trained Qwen and Llama models substantially outperform the GPT-4.1 few-shot baseline.
Despite the same underlying questions, overall performance in the 2Wiki-Text environment is notably lower than in 2Wiki-KG.
This disparity can be attributed to the differing data modalities and interaction paradigms.
Interacting with KG may more closely resemble navigating discrete choices (e.g., selecting entities or relations), potentially reducing the burden on complex retrieval or synthesis from noisy search results.
Conversely, text-based interaction requires robust open-retrieval over the corpus where explicit relationships between retrieved passages are often absent, demanding more sophisticated synthesis capabilities.
Nevertheless, the consistent improvements observed affirm MyGO's efficacy across these varied settings.

\begin{table}[h] 
    \centering 
    \renewcommand{\arraystretch}{1.1} 
    \caption{Comparative Performance of Qwen on Hotpot and Hotpot-Kimi Environment.}
    \label{tab:hotpot}
    \resizebox{0.46\textwidth}{!}{
    \begin{tabular}{@{}llcccc@{}} 
    \toprule
    \textbf{Type} & \textbf{Model} & \multicolumn{2}{c}{\textbf{Hotpot}} & \multicolumn{2}{c}{\textbf{Hotpot-Kimi}} \\
    \cmidrule(lr){3-4} \cmidrule(lr){5-6} 
     &  & \textbf{EM} & \textbf{F1} & \textbf{EM} & \textbf{F1} \\
    \midrule
    \multirow{1}{*}{Few-Shot} & GPT-4.1 &  29.80&  48.09&  33.50&  53.35\\
    \midrule
    Few-Shot & \multirow{3}{*}{\makecell{Qwen2.5-7B}}& 15.77 & 27.96 & 15.60 & 32.70 \\
    WarmUp &   & 40.55 &  52.35 & 52.51 & 66.04 \\
    MyGO &  & 41.54   &  53.79 & 54.07 & 68.48 \\
    \bottomrule
    \end{tabular}
    }
\end{table}

\begin{table*}[h!]
\centering
\caption{End-to-end textual multi-hop question answering results. 
}
\begin{tabular}{lllcccccc}
\toprule
\multicolumn{1}{c}{\multirow{2}{*}{\textbf{Method}}} & \multicolumn{1}{c}{\multirow{2}{*}{\textbf{Backbone}}} & \multicolumn{1}{c}{\multirow{2}{*}{\textbf{Retriever}}} & \multicolumn{2}{c}{\textbf{HotpotQA}}  & \multicolumn{2}{c}{\textbf{MuSiQue}}   & \multicolumn{2}{c}{\textbf{2Wiki-Text}}      \\
\cmidrule(lr){4-5} \cmidrule(lr){6-7} \cmidrule(lr){8-9} 
\multicolumn{1}{c}{}                        & \multicolumn{1}{c}{}                          & \multicolumn{1}{c}{}                           & \textbf{EM}            & \textbf{F1}            & \textbf{EM}            & \textbf{F1}            & \textbf{EM}             & \textbf{F1}            \\
\midrule
Direct Inference                            & Qwen 2.5-7B                                   & - & 18.3 & - & 3.1 & - & 25.0 & - \\
CoT & Qwen 2.5-7B & - & 9.2 & - & 2.2 & - & 11.1 & - \\
RAG                                         & Qwen 2.5-7B                                   & E5                                     &  29.9  & -                               & 5.8    & -                          & 23.5      & -                         \\
RAFT                                       & Llama 3.1-8B                                  & Dragon-Plus                                    & 41.0                      & 51.6                            & 13.8                       & 24.0                       & 39.4                          & 45.8                      \\
RaFe                                        & GPT 4o-mini                                   & Google Search                                  & 40.6                      & 55.4                            & 11.8                       & 23.8                       & 36.2                          & 39.3                      \\
HippoRAG                                    & GPT 3.5-turbo                                 & ColBERTv2                                      & \underline{45.7}                & \textbf{59.2}                   & 21.9                       & 33.3                       & 47.7                          & \underline{62.7}                      \\
Iter-RetGen                                 & Qwen 2.5-7B                                   & FlashRAG                                       & 34.4  & -                               & 8.7                        & -                          & 27.9                          & -                         \\
IRCoT                                       & Qwen 2.5-7B                                   & BM25                                           & 30.3  & -                               & 7.0                        & -                          & 21.6                          & -                         \\
IterDRAG                                    & Gemini 1.5                                    & Gecko 1B                                       & 38.4                & 49.8                            & \underline{22.6}                       & 35.0                       & 44.3                          & 54.6                      \\
RAG-Star                                    & Llama 3.1-8B                                    & bge-large-en-v1.5                                       & 42.0                & 54.4                            & 13.0                       & 22.2                       & 34.0                          & 42.0                      \\
\midrule
\textit{RAG Agents} & & & & & & & & \\
Search-o1                                   & Qwen 2.5-7B                                       & E5                                & 18.7                      & -                      & 5.8                       & -                       & 17.6                    & -             \\
Search-o1                                   & QwQ 32B                                       & Bing Web Search                                & 45.2                      & \underline{57.3}                      & 16.6                       & 28.2                       & 58.0                    & \textbf{71.4}             \\
Search-R1 & Qwen 2.5-7B & E5 & 37.0 & - & 14.6 & - & 41.4 & - \\
RAG-Gym                                     & Llama 3.1-8B                                  & bge-base + BM25                                & 44.1                      & 56.8                           & -                          & -                          & 50.2                          & 57.9            \\
ReSearch                                    & Qwen 2.5-7B                                   & FlashRAG                                     & 43.5                      & -                               & 22.3                       & -                          & 47.6                          & -                         \\
\midrule
\textit{Settings}         & & & \multicolumn{2}{c}{\textit{In-domain}} & \multicolumn{2}{c}{\textit{Out-of-domain}} & \multicolumn{2}{c}{\textit{In-domain}}  \\ 
\textbf{Mujica (few-shot)}                  & Qwen 2.5-7B                                   & bge-large-en-v1.5                              & 15.77 & 27.96 & 5.34   & 17.81  & 23.35    & 33.52 \\
\textbf{Mujica w/ warm up}                      & Qwen 2.5-7B                                   & bge-large-en-v1.5                              & 40.55                     & 52.35                           & N/A                        & N/A                        & \underline{50.18}                         & 56.25                     \\
\textbf{Mujica-MyGO}                        & Qwen 2.5-7B                                   & bge-large-en-v1.5                              & 41.54                     & 53.79                           & \textbf{26.11}             & \textbf{35.92}             & \textbf{53.17}                & 59.62               \\
\bottomrule
\end{tabular}
\label{exp_baselines}
\end{table*}


Table~\ref{tab:hotpot} details the results on the HotpotQA dataset.
Notably, even the WarmUp models exhibit strong performance, outperforming the few-shot GPT-4.1 baseline.
However, the performance gain from MyGO over the WarmUp models is less pronounced in the Hotpot environment.
We attribute this observation to two limitations inherent in this setup.
Firstly, the efficacy of the dense retriever can be a bottleneck;
essential evidence may not be surfaced for some queries.
Secondly, the HotpotQA evaluation protocol has no answer aliases.
This is problematic because LLMs frequently generate answers that are semantically equivalent to the ground truth but lexically different, which are penalized by strict string-matching metrics.
To mitigate the impact of imperfect retrieval, we further evaluate using the Hotpot-Kimi environment, which is designed to simulate near-perfect retrieval by providing gold supporting passages.
In this idealized Hotpot-Kimi setting, MyGO demonstrates substantial and consistent performance improvements over the WarmUp models, underscoring its effectiveness when reasoning capabilities can be more directly assessed.

\subsubsection{\textbf{Effectiveness of Mujica}}
In order to evaluate the contribution of our two critical design choices in the proposed Mujica framework, we replace the planner–worker split with vanilla iterative RAG, and substitute DAG-like reasoning with chain-like reasoning. For the sake of fair comparison, all subsequent experiments are conducted with a pre-trained Qwen 2.5-7B-Instruct model, which eliminates confounding effects introduced by model training.

\begin{table}[h]
\centering 
\renewcommand{\arraystretch}{1.1} 

\caption{Experimental results for design choices of the proposed Mujica framework.}
\label{tab:mujica}
\resizebox{0.46\textwidth}{!}{
\begin{tabular}{@{}lcccc@{}} 
\toprule
\textbf{Settings} & \multicolumn{2}{c}{\textbf{2Wiki-KG}} & \multicolumn{2}{c}{\textbf{2Wiki-Text}} \\
\cmidrule(lr){2-3} \cmidrule(lr){4-5} 
 &  \textbf{EM} & \textbf{F1} & \textbf{EM} & \textbf{F1} \\
\midrule
Mujica (Qwen2.5-7B) & 57.60 & 62.24 & 23.35 & 33.52 \\
- w/o. planner-worker split & 54.85 & 59.63 & 23.22 & 33.40 \\
- w/o. DAG-like reasoning & 53.75 & 59.08 & 21.73 & 32.36 \\
\bottomrule
\end{tabular}
}
\end{table}
From Table~\ref{tab:mujica} we conclude that our proposed DAG-like reasoning paradigm not only enables independent sub-questions to be processed in parallel, but also outperforms the chain-style baseline. We attribute the performance gain to the paradigm’s ability to effectively decompose complex questions and resolve sub-problems through optimal reasoning paths. The structured decomposition allows the model to better capture the underlying logic of multi-hop questions, leading to more accurate and robust reasoning outcomes. In addition, our planner–worker architecture surpasses the vanilla iterative RAG setup. We attribute the better performance to Mujica's ability to prevent irrelevant information from interfering with the model's reasoning and planning process.

\subsubsection{\textbf{Mujica-MyGO v.s. Baseline Methods}}
\label{appendix_comparision}
In order to comprehensively evaluate the performance of our proposed solution, we compare Mujica-MyGO with a series of RAG systems. Apart from direct inference with pretrained LLM~\cite{Qwen2} and vanilla RAG~\cite{RAG}, we consider 7 iterative RAG solutions: RAFT~\cite{zhang2024raftadaptinglanguagemodel}, RAFe~\cite{RaFe}, HippoRAG~\cite{gutiérrez2024hipporag}, Iter-RetGen~\cite{Iter-RetRAG}, IRCoT~\cite{IRCoT}, IterDRAG~\cite{IterDRAG}, RAG-star~\cite{RAG-Star} and 4 RAG agents: Search-o1~\cite{Search-o1}, RAG-Gym~\cite{RAG} and ReSearch~\cite{ReSearch}. 
Considering the differences in backbone LLMs and retrieval methods employed by each RAG system, it is inappropriate to compare the performance of different RAG systems solely based on their evaluation metrics.

Table~\ref{exp_baselines} presents the experimental results on three representative multi-hop question answering datasets, which shows that \textit{Mujica-MyGO is an effective and efficient multi-hop QA solution}. Despite leveraging a 7B Qwen2.5 LLM as its backbone, Mujica-MyGO achieves state-of-the-art performance on the\allowbreak MuSiQue and 2Wiki datasets and delivers competitive results on HotpotQA. 
One may notice that some baseline methods outperform Mujica-MyGO on the HotpotQA dataset. However, these approaches typically rely on larger-scale backbone LLMs and/or more sophisticated retrieval methods. In general, larger backbone models (e.g., GPT and QwQ) tend to yield better results. 
In addition, incorporating advanced retrievers (e.g., HopRetriever~\cite{HopRetriever}, Dragon-plus~\cite{DRAGON}) or expanding the knowledge corpus via search engines (e.g., Google, Bing) can further enhance performance. Nevertheless, it is important to emphasize that improving retrieval is not the focus of this paper. In order to evaluate the effectiveness of the proposed method, we opt to adopt the official corpus released with the HotpotQA dataset~\cite{yang2018hotpotqa}. Across model variants, we find that even without costly reinforcement learning, a small amount of SFT data is sufficient for Mujica to surpass baseline methods based on question decomposition and iterative retrieval, the method to consistently surpass, demonstrating its robustness. However, every approach has its trade-offs. As a complex question-answering pipeline, Mujica imposes substantial demands on the LLM's instruction-following and reasoning capabilities, making it difficult for small-scale pre-trained LLMs to achieve desirable results in few-shot settings.

\subsection{Additional Studies}
\subsubsection{\textbf{Effectiveness in Out-Of-Domain Settings}} 
To assess generalization capabilities, we conduct out-of-distribution (OOD) evaluations where models are RL post-trained on one dataset and evaluated on another one.
Specifically, we examine cross-dataset transfer from HotpotQA to Musique (text-based), and from 2Wiki-KG to Qald-10 (KG-based).
As shown in Table~\ref{tab:ood}, MyGO demonstrates strong OOD performance.
As presented in Table~\ref{tab:ood}, models trained with MyGO demonstrate robust OOD generalization.
Specifically, applying MyGO on the source datasets leads to significant performance improvements on the respective target OOD datasets when compared to the few-shot baseline.
These findings indicate that MyGO does not overfit to the training data; rather, it appears to foster the acquisition of more generalizable reasoning and decision-making strategies.
This capacity for effective generalization to unseen datasets aligns with a primary motivation for employing RL in the development of intelligent agents.
\begin{table}[h]
\label{table_2}
    \centering 
    \renewcommand{\arraystretch}{1.1} 
    \caption{Out-of-distribution Evaluation on Qwen 2.5-7B: HotpotQA$\to$Musique and 2Wiki-KG$\to$QALD-10.}
    \vspace{-0.5em}
    \label{tab:ood}
    \vspace{.5em}
    \resizebox{0.45\textwidth}{!}{
    \begin{tabular}{@{}llcccc@{}} 
    \toprule
    \textbf{Type} & \textbf{Model} & \multicolumn{2}{c}{\textbf{Hotpot} $\to$ \textbf{MuSiQue}} & \multicolumn{2}{c}{\textbf{2Wiki} $\to$ \textbf{QALD}} \\
    \cmidrule(lr){3-4} \cmidrule(lr){5-6} 
     &  & \multicolumn{1}{c}{\textbf{EM}} & \multicolumn{1}{c}{\textbf{F1}} & \multicolumn{1}{c}{\textbf{EM}} & \multicolumn{1}{c}{\textbf{F1}} \\
    \midrule
    Few-Shot & \multirow{2}{*}{\makecell{Qwen\\2.5-7B}}& 5.34& 17.81& 27.92& 37.11\\
    MyGO &  & 26.11&  35.92& 39.85& 49.73\\
    \bottomrule
    \end{tabular}
    }
\end{table}

\subsubsection{\textbf{MyGO v.s. Other Proximity Optimization Approaches}} 
Compared with ReSearch (using GRPO), Search-R1 and RAG-Gym (using PPO), Mujica-MyGO achieves comparable or even better performance across three datasets with the commonly adopted bge-large-en-v1.5 dense retriever~(see Table~\ref{po_solutions}). This demonstrates that MyGO simplifies GRPO and PPO without compromising performance on multi-hop QA tasks. Moreover, MyGO does not require additional resources to hold a reference model or to compute advantages, striking a great balance between efficiency and performance. 


\begin{table}[h!]
\centering
\caption{Comparison of Proximity Optimization-based Multi-hop QA Solutions.
}
    \vspace{-0.5em}
\begin{tabular}{lcccc}
\toprule
\multicolumn{1}{c}{\textbf{Pipeline}} & \textbf{P.O. Method} & \textbf{HotpotQA} & \textbf{MuSiQue} & \textbf{2Wiki} \\
\midrule
Search-R1                    & PPO                  & 37.0     & 14.6    & 41.4  \\
RAG-Gym                      & PPO                  & \textbf{44.1}     & -       & 50.2  \\
ReSearch                     & GRPO                 & 43.5     & 22.3    & 47.6  \\
\midrule
\textbf{Mujica}                       & \textbf{MyGO}                 & 41.54    & \textbf{26.11}   & \textbf{53.17} \\
\bottomrule
\end{tabular}
\label{po_solutions}
\end{table}

\subsubsection{\textbf{Runtime Comparison - MyGO v.s. GRPO}}

We further present a runtime comparison between MyGO and GRPO.
As shown in Table~\ref{tab:grpo_vs_grpo}, Mujica trained with MyGO achieves performance comparable to that obtained with GRPO, while requiring approximately 74\% less training time and about 94\% fewer training trajectories.


\begin{table}[h!]
\centering
\caption{Comparison to GRPO. \# of t.d.: means the number of trajectories for the planner and the worker; t. time: denotes the total training time.}
\label{tab:grpo_vs_grpo}
\begin{tabular}{lcccc}
\hline
Algo. & F1 & EM & \# of t.d. & t. time\\
\hline
GRPO & 86.06 & 80.15 & 137,941 & $\sim$135min \\
MyGO & 85.70 & 80.30 & 7,343 & $\sim$35min \\
\hline
\end{tabular}
\end{table}



\section{Related Work}
\paragraph{MHQA with textual corpus.}
Retrieval-based methods dominate the landscape of textual MHQA since supporting contexts from external corpora can effectively compensate for deficiencies in the inherent knowledge of language models and mitigate hallucinations. Traditional approaches primarily utilized sentence transformers~\cite{bgem3} for dense retrieval or BM25~\cite{BM25} for keyword matching. Recently, extensive research has focused on optimizing the QA pipeline, particularly through the decomposition of complex questions~\cite{IM-RAG,ReAct}, query rewriting and refinement~\cite{RQ-RAG,RaFe}, and the iterative retrieval of relevant evidences in multiple steps~\cite{SelfRAG,ReSP,Iter-RetRAG,IRCoT}. 
\paragraph{Knowledge-based question answering.}
Early approaches treat KBQA as a semantic parsing task, which generates and executes SPARQL queries~\cite{lan-jiang-2020-query,SPARQL2021,KD-CoT}. These methods usually suffer from syntax errors and inaccurate entity labels, resulting in limited accuracy. Recently, researchers have shifted towards  leveraging RAG and the advanced semantic understanding capabilities of LLMs to generate answers~\cite{RRA}. LLM-based KBQA approaches can be generally categorized into two types: iterative exploration and subgraph reasoning. Iterative exploration methods~\cite{ToG,ToG2,ToG-I,KG-CoT,PoG,FiDeLiS,GoG,KARPA,Paths-over-Graph,BeamRetrieval,DoG} progressively retrieve and verify KG elements through multi-step reasoning, balancing interpretability with computational overhead. These approaches rely heavily on LLMs' inherent reasoning capabilities due to the lack of supervised intermediate steps in most datasets. In contrast, Subgraph reasoning methods~\cite{RoG,GNN-RAG,SubgraphRAG,READS,GCR} extract relevant KG substructures in a single retrieval step for LLM-based answering, enabling efficient fine-tuning but suffering from the incompleteness of retrieved subgraphs and underlying knowledge bases.

\section{Conclusion}
In this work, we introduce Mujica, an interactive multi-agent framework for RAG systems, alongside MyGO, a novel reinforcement learning method designed to simplify its optimization.
Mujica employs a planner-worker architecture to systematically decompose and resolve complex questions, while MyGO enhances training efficiency through an intelligent sampling strategy.
Our experiments demonstrate that the proposed framework not only achieves significant performance improvement across multiple QA datasets but also exhibits strong inductive reasoning.
These results underscore its scalability and potential to generalize effectively in real-world scenarios.

\newpage






\bibliographystyle{ACM-Reference-Format} 
\bibliography{ref}


\appendix
\onecolumn

\section{Problem Definition}
\label{probelm_definition}
The MHQA task involves answering a complex question $q$ by performing multi-step reasoning over multiple pieces of supporting contexts $\mathcal{C}=\{c_1, c_2, \cdots, c_{|\mathcal{C}|}\}$ to derive the final answer $A_q$. Based on the form of supporting contexts, MHQA can be further breakdown into two sub-tasks, text-based MHQA and Knowledge-based Question Answering~(KBQA).

\paragraph{Text-based Multi-hop Question Answering}
In text-based MHQA, the supporting contexts are paragraphs scatter in one single document or multiple passages. In practice, the text-based MHQA task has two different settings. The \textit{distractor} setting provides a pre-extracted set of supporting paragraphs $\mathcal{T}_q$ for each question $q$, which includes both relevant information and distractors. In contrast, under the \textit{full-corpus} setting, all questions share an open textual corpus $\mathcal{T}$ (e.g. Wikipedia). In this work, we focus on the \textit{full-corpus} setting, since it is closer to the real-world scenario. 

\paragraph{Knowledge-based Question Answering}
Under the setting of KBQA, the supporting contexts are retrieved from a knowledge graph~(KG). KGs consist a large set of entities~$\mathcal{E}$ and their relations~$\mathcal{R}$, typically represents as triples $\mathcal{T}$ in the form of $(e_h,r,e_t)$, where $e_h$ and $e_t$ denote the head and the tail entity, respectively, and $r$ is the relation connecting them. To answer a question $q$, the model needs to identify the topic entities mentioned in the question, and traverse the KG by chaining multiple triples together to obtain the final answer $A_q$.

\section{Mathematical Proofs}
\label{appendix_math_proof}
\klapprox*
\begin{proof}
Recall that $\pi^*(\tau) = \frac{\exp(r(\tau)/\alpha)}{Z(\alpha)}$ and $Z(\alpha) = \int \exp(r(\tau)/\alpha) d\tau$, we have the following statement for $\pi^{>K}$:
\begin{equation}
    \pi^{>K}(\tau) = \frac{\exp (r(\tau)/\alpha)}{Z^{>K}(\alpha)} \; , \quad \text{where } Z^{>K}(\alpha) = \int_{\mathcal{T}} \mathds{1}(r>K) \exp (r(\tau')/\alpha) d\tau' \; ,
\end{equation}
where $\mathds{1}(r>K)$ is the characteristic function filtering out trajectories whose reward is less than $K$.
The above statement tells that $\pi^{>K}$ is actually a conditional probability of $\pi^*$, i.e., $\pi^{>K}(\tau)=\pi(\tau | r(\tau)>K)$.
Also, the above statement allows us to compute the KL divergence easily:
\begin{equation}
    \mathds{D}_{\text{KL}}(\pi^{>K} || \pi^*) = \log \frac{Z}{Z^{>K}} \; .
\end{equation}
Notably,
\begin{equation}
     Z^{>K} = \int_{r>K} \exp(r / \alpha) dr = (M-K)\exp(r'/\alpha) \; ,
\end{equation}
where $M = \sup (r) $ and the last equality is obtained by the mean value theorem.
In summary, if $K < M - \frac{Z}{\exp(\delta) \exp(r'/\alpha)}$, then the proposition holds.

\end{proof}

\varapprox*
\begin{proof}
First, we expand the variance by regarding $\pi^{>K}$ as a distribution with respect to $r$.
Note that all $\Var$ and $\E$ here are on $\pi^{>K}$ and $M=\sup(r)$.
\begin{equation}
    \Var[r] = \E \left[r - \E \left[ r \right] \right]^2 = \E \left[r^2 \right] - \E^2 \left[r \right] 
    =\frac{\int^M_K r^2 \exp(r/\alpha) dr}{ \int^M_K \exp(r/\alpha)dr} - \left( \frac{\int^M_K r \exp(r/\alpha) }{\int^M_K \exp(r/\alpha)} \right)^2 \; .
\end{equation}
Note that we can expand the integral at $\alpha=0$ with a Taylor series:
\begin{equation}
    \begin{aligned}
    \Var[r] =
    \frac{e^{(K+M)/\alpha}(-2\alpha^2 + O(\alpha^6)) + \left( e^{(2K)/\alpha} + e^{(2M/\alpha)}\right)\left(\alpha^2 + O(\alpha^6) \right) + (K-M)^2(-e^{(K+M)/\alpha}) }{\left(e^{(K/\alpha)} - e^{(M/\alpha)} \right)^2} \; .
    \end{aligned}
\end{equation}
Notice the fact that $Z^{>K}(\alpha)=\alpha(\exp(M/\alpha) - \exp(K/\alpha))$, the proposition holds.
\end{proof}

 \section{Datasets and Metrics}
 \label{appendix_datasets}
To evaluate MyGO's effectiveness on complex question answering, we employ four representative datasets: 2WikiMultihopQA (2Wiki)~\cite{ho2020constructing}, QALD-10~\cite{QALD-10}, HotpotQA~\cite{yang2018hotpotqa}, and MuSiQue~\cite{trivedi2021musique}.
2Wiki utilizes both structured (Wikidata) and unstructured (Wikipedia) data to create multi-hop questions.
QALD-10 was initially designed for Multilingual Question Answering over KG but late was widely used as a benchmark on KBQA \cite{ToG}.
The other two datasets focus on multi-document and multi-hop reasoning. HotpotQA features complex questions requiring reasoning across multiple documents. MuSiQue consists of questions demanding two to four reasoning hops, constructed from five existing single-hop QA datasets.

\section{Implementation Details}
\label{appendix_implementation_detail}
\subsection{Sampling Strategies}

\paragraph{Reward Criterion}

Ideally, we would use the ``correctness'' of predicted answers as the reward criterion.
However, this ``correctness'' is not directly accessible in practice.
The ground truths provided in QA task datasets are in fact the golden answers.
The failure to achieve an exact match - a widely used metric in QA tasks - between the generated answer and the golden answer does not necessarily indicate the semantic ``incorrectness''.
For example, for a question in HotpotQA ``\emph{John T. Mather Memorial Hospital is located in a village in what Suffolk County, New York town?}'', 
the generated answer ``Brookhaven'' fails to achieve an exact match with the golden answer ``Town of Brookhaven'', which would be considered correct by a human judge.

Therefore, we adopt the F1 score as our evaluation criterion
 and consider a generated answer correct if it achieves a sufficiently high F1 score close to 1.

Note that some works utilize more complicated non-standard criteria, e.g., LLM-as-Judge~\cite{ReSearch,sun2025zerosearch}, Cover Exact Match (CEM)~\cite{sun2025zerosearch, song2025r1}.
Since this is not the main focus of this work, we simply follow the standard criterion widely used in previous works.

\paragraph{Progressive Selection Threshold}

As stated in section~\ref{sec:mygo} as well as the aforementioned, we aim to directly sample trajectories $\tau$ for $r(\tau) > K$ with a sufficiently large $K$.

However, this might lead to very low sample efficiency when the current policy is not sufficiently strong or when the tasks are particularly challenging. As a compromise, we consider a progressive selection threshold defined as $K:= \max(K', (\bar{r}_n/(r_\text{sup} + 1))\cdot r_\text{sup})$, where $r_\text{sup}$ denotes $\sup_{\tau \in \mathcal{T}} r(\tau)$ and $\bar{r}_n$ represents the empirical average reward of the trajectories for a batch of $n$ questions at the first pass. 
This approach balances both the difficulty of sampling satisfactory trajectories and the quality of the sampled trajectories. As the policy improves during the MyGO training procedure, $K$ will asymptotically grow to an acceptable value, thereby satisfying Proposition~\ref{propos_1}.

\paragraph{Multiple-Attempt Sampling}

For a batch of $n$ questions, 
we would iteratively sample trajectories for them.
For a question, the sampling procedure will stop if $m$ satisfying trajectories are sampled or reach the predetermined maximum sampling attempts.
With vLLM engine~\cite{kwon2023efficient}, we could accelerate the sampling procedures in parallel efficiently with multi-threading/multi-processing.

\subsection{Training}

\paragraph{Online Off-policy Optimization}

In MyGO, we primarily focus on online off-policy optimization. 
Given a total of $N$ data examples (i.e., questions), we divide them into $T$ batches for $T$ update iterations.
For each iteration, we sample trajectories (including answers) from $N/T$ questions to serve as pseudo-labels and optimize the models via MLE through a standard SFT pipeline (e.g., LLaMA-Factory~\cite{zheng2024llamafactory}).
After completing one update iteration, we proceed to the next sampling iteration.

Notably, due to the off-policy strategy, 
we can parallelize the sampling and training processes after the first iteration, when a sufficient number of GPUs are available.

\paragraph{Trajectory Selection for Training.}
In the Mujica framework, the planner and workers collaborate to reach the final answer. This interaction generates a single planner trajectory and multiple worker trajectories. It is important to clarify that in Section~\ref{sec:mygo}, the term "trajectory" specifically refers to the planner's trajectory. In contrast, the dialogue history for each worker is straightforward, comprising a single-turn interaction with the environment. The environment's message to the worker consists of the question from the planner and passages from the retriever, while the worker's message contains its response. If the reward associated with a planner trajectory exceeds a predefined threshold, that trajectory and its corresponding worker trajectories are selected as independent data for training.

\paragraph{Training Advances from Pre-training/SFT.}
Unlike other reinforcement learning techniques, MyGO's training process can be framed as supervised fine-tuning (SFT) using self-generated data and labels. Consequently, we can directly leverage existing training advancements from pre-training and SFT techniques, including optimizers, learning rate schedulers, and hyperparameter settings.

\subsection{Hyperparameters and Configurations}

\paragraph{Sampling.}
For all environments and datasets, we sample a maximum of $m=3$ trajectories per question from a randomly selected subset of 20,000 questions from the training sets. The maximum number of sampling attempts is set to 16. During each iteration of the sampling process, we sample trajectories from 1,000 questions.

\paragraph{Warmup.}
To burn in the prompts, we introduce a warmup stage before performing MyGO, in which we use the few-shot model to randomly sample 300 trajectories with $\text{EM}=1$ from the training sets. We then fine-tune the pre-trained model with these trajectories following the same procedure as MyGO.

\paragraph{Configuration.}

Following the settings of regular SFT, we employ the AdamW optimizer~\cite{loshchilov2019decoupled} with a Cosine Annealing learning rate scheduler~\cite{loshchilov2017sgdr}. For the warmup phase, we configure the initial learning rate at 1e-5 and the minimum learning rate at 1e-6. In the MyGO training procedure, we set the initial learning rate to 5e-7 and the minimum learning rate to 1e-8.
We leverage vLLM~\cite{kwon2023efficient} as our inference engine and LLaMA-Factory~\cite{zheng2024llamafactory} as our training framework.

\section{Further Study: Offline MyGO?}
\label{appendix_offline}
Designing an offline variant of MyGO is straightforward: we can use the initial policy after warmup to sample all required trajectories. In this work, we present a preliminary attempt using Qwen on the 2Wiki-KG and 2Wiki-Text datasets.

\begin{figure}[htbp]
\centering
\begin{minipage}{0.48\linewidth}
\centering
\includegraphics*[width=\linewidth]{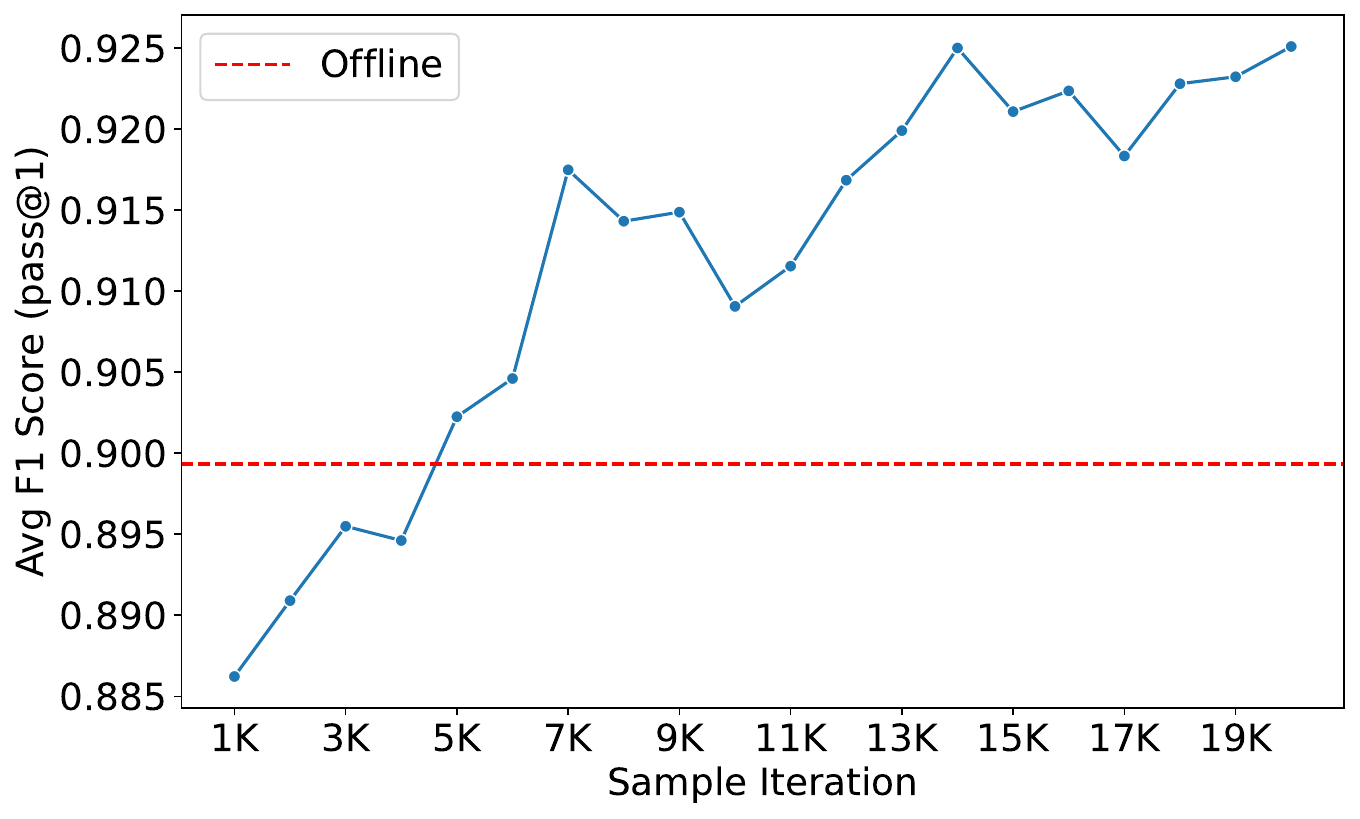}
\caption{F1 score on 2Wiki-KG training}
\label{fig:first}
\end{minipage}%
\hfill%
\begin{minipage}{0.48\linewidth}
\centering
\includegraphics*[width=\linewidth]{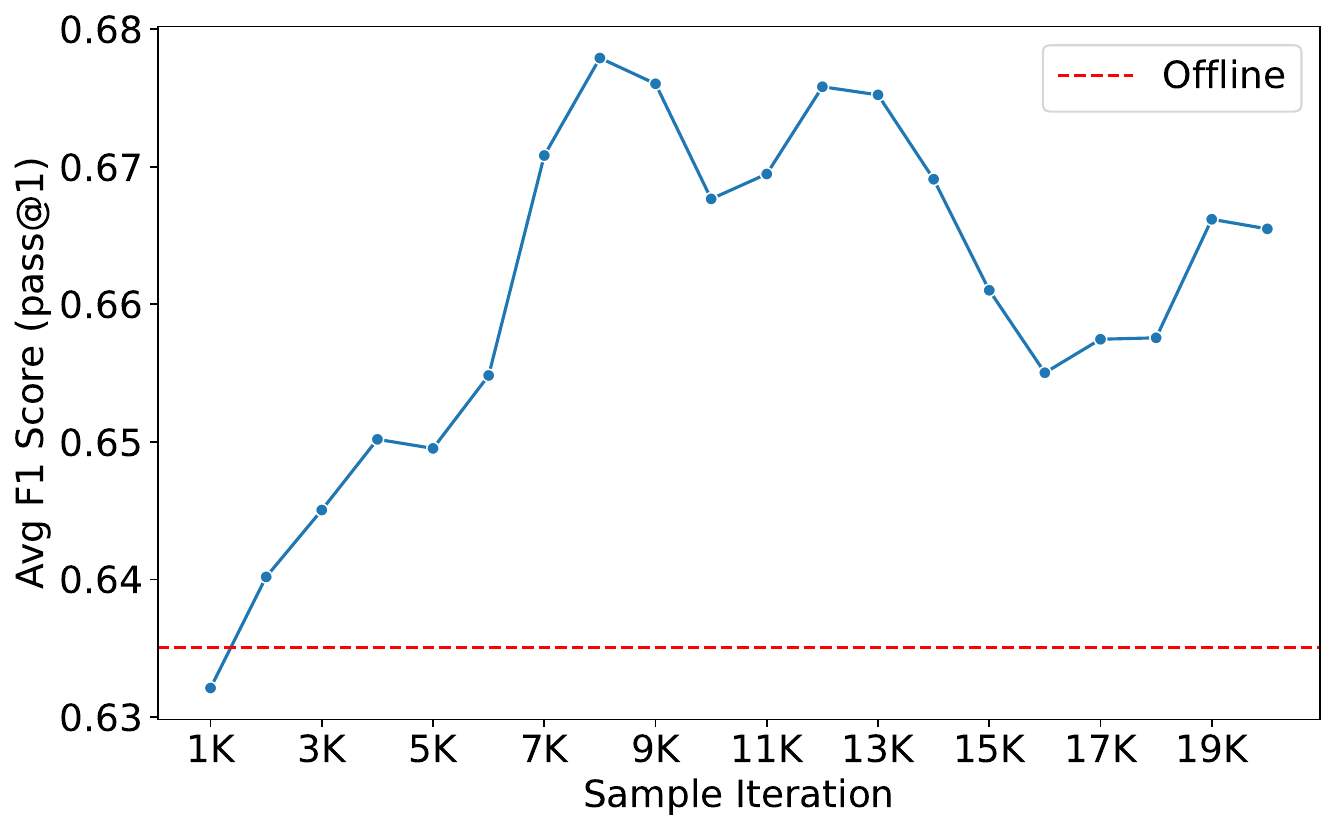}
\caption{F1 score on 2Wiki-Text training}
\label{fig:second}
\end{minipage}
\label{fig:combined}
\end{figure}



\begin{table}[h!]

\label{tab:online_vs_offline}
\centering 
\renewcommand{\arraystretch}{1.1} 
\caption{Mujica+MyGO w/ Qwen  Online v.s. Offline on 2Wiki-KG and 2Wiki-Text Environments.}
\vspace{0.2cm}
\label{tab:model_online_offline}

\begin{tabular}{@{}llcccc@{}} 
\toprule
\textbf{Type} & \textbf{Model} & \multicolumn{2}{c}{\textbf{2Wiki-KG}} & \multicolumn{2}{c}{\textbf{2Wiki-Text}} \\
 
\cmidrule(lr){3-4} \cmidrule(lr){5-6} 
\textbf{Mujica+} &  & \textbf{EM} & \textbf{F1} & \textbf{EM} & \textbf{F1} \\
\midrule
\multirow{1}{*}{Few-Shot} & GPT-4.1 & 78.50 & 84.24 & 30.70  & 52.14 \\
\midrule
Few-Shot & \multirow{4}{*}{Qwen2.5-7B}& 57.49 & 62.31 & 12.41 & 19.71 \\
WarmUp &  & 74.93 & 80.74 & 50.18 & 56.25 \\
\cmidrule{0-0} 
\cmidrule{3-6} 
MyGO-Offline &  & 77.09 & 83.16 & 53.27 & 59.78 \\
MyGO-Online &  & 77.63 & 84.15 & 53.17 & 59.62 \\
\bottomrule
\end{tabular}

\end{table}

We note that our goal is not to compare online and offline RL, 
as this has been discussed in many prior works~\cite{tang2024understanding}.
Instead, we just aim to present an alternative approach for implementing MyGO.

From Proposition~\ref{propos_1}, we can expect that the quality of samples obtained from MyGO-offline and MyGO-online should be comparable, as both methods sample directly from an asymptotically optimal policy.
The former uses a fixed threshold $K$, while the latter can dynamically and progressively increase $K$, which aligns more closely with Proposition~\ref{propos_1}.
In the demonstration experiments (as shown in Table~\ref{tab:model_online_offline}), we observe that MyGO-Offline can also achieve outstanding performance, being slightly worse or comparable to MyGO-Online.
This indirectly justifies the insight from Proposition~\ref{propos_1}.

\section{Extended Related Works}
\paragraph{MHQA with textual corpus.}
Retrieval-based methods dominate the landscape of textual MHQA since supporting contexts from external corpora can effectively compensate for deficiencies in the inherent knowledge of language models and mitigate hallucinations. Traditional approaches primarily utilized sentence transformers~\cite{bgem3} for dense retrieval or BM25~\cite{BM25} for keyword matching. Recently, extensive research has focused on optimizing the QA pipeline, particularly through the decomposition of complex questions~\cite{IM-RAG,ReAct} and the iterative retrieval of relevant evidences in multiple steps~\cite{SelfRAG,ReSP,Iter-RetRAG,IRCoT}. Most notably, ReAct~\cite{ReAct} integrates retrieval actions into the reasoning trajectory exploration process, enabling dynamic interaction with external environments. IM-RAG~\cite{IM-RAG} optimizes multi-round retrieval and reasoning via reinforced learning. Search-o1~\cite{Search-o1} introduces an agentic RAG framework, empowering the language model to decide when to retrieve external knowledge and to refine retrieved contexts, achieving state-of-the-art performance.

\paragraph{Knowledge-based question answering.} 
Early approaches~\cite{lan-jiang-2020-query,SPARQL2021,KD-CoT} formulates KBQA as a semantic parsing task, focusing on extracting entities and relations from natural language questions and translating them into SPARQL statements capable of executing logical queries over KGs~\cite{TIARA,SymKGQA}. While these methods provide accurate and interpretable answers, they encountered significant limitations since SPARQL statements generated by language models often suffer from syntactic errors, including incorrect entity and relation labels as well as unexpected grammatical mistakes~\cite{DecAF,RoG,DoG}. Moreover, the incomplete nature of KGs~\cite{KGR3} can also lead to query failures, preventing these approaches from outputting correct answers. 

Recently, researchers have shifted towards treating KBQA as an information retrieval~(IR) task, leveraging RAG and the advanced semantic understanding capabilities of LLMs to generate answers based on factual knowledge stored in KGs~\cite{RRA}. LLM-based KBQA methods can be generally categorized into two types: iterative exploration and subgraph reasoning. Iterative exploration methods~\cite{ToG,ToG2,ToG-I,KG-CoT,PoG,FiDeLiS,GoG,KARPA,Paths-over-Graph,BeamRetrieval,DoG} adopt a beam search-like strategy to retrieve the most relevant relations and neighboring entities, followed by pruning operations guided by LLMs. These methods exhibit strong traceability, enabling responsible and interpretable reasoning, but often come with the drawback of high computational complexity~\cite{FastToG}. Moreover, the absence of ground truth reasoning paths and intermediate nodes in most KBQA datasets limits the feasibility of applying SFT~\cite{RoG,GNN-RAG}. As a result, their performance is inevitably constrained by the inherent reasoning capabilities of LLMs. 
In contrast, subgraph reasoning methods~\cite{RoG,GNN-RAG,SubgraphRAG,READS,GCR} retrieve all candidate reasoning paths related to the question from the local neighborhood of the topic entities and guide the LLM to generate answers based on these paths in a single step. This approach enable direct improvement of the LLM's in-context learning capabilities through SFT. Nevertheless, its effectiveness is often constrained by the quality of retrieved context, which typically cover only a subset of the knowledge required to address complex questions, omitting other critical information~\cite{PoG}. Furthermore, due to the incompleteness of KGs, the existence of relevant reasoning paths within the subgraph is not guaranteed. In such scenarios, LLM may hallucinate and produce incorrect answers.

\paragraph{Simplifying Reinforcement Learning for LLM postraining.}
Recent research has focused on simplifying RL algorithms for LLMs.
For instance, GRPO~\cite{shao2024deepseekmath} simplifies PPO by replacing the value model with standard deviation.
Similarly, ReMax~\cite{li2024remax} adapts VPO by using the logits of greedy decoding as a baseline for GAE.
Other VPO modifications include RLOO~\cite{ahmadian2024back}, REINFORCE++\cite{hu2025reinforce++}, and GPG~\cite{chu2025gpg}.
Another line of research replaces VPO with MLE through advanced sampling techniques. Within this category, RAFT~\cite{zhang2024raftadaptinglanguagemodel} selects the highest-reward trajectory within a group, while RAFT++~\cite{xiong2025minimalist} enhances this by incorporating rejection sampling. Both RAFT and RAFT++ can be viewed as specific instances of STaR~\cite{zelikman2022star}.
MyGO aligns with this latter research direction. It selects trajectories using either a fixed or progressively improving threshold, a property ensuring that sampling approaches the optimal policy. We further provide a mathematical theory to explain the efficacy of this streamlined method.
If we continue to simplify MyGO, it will converge to STaR~\cite{zelikman2022star}.
However, our primary motivation stems from the goal of simplifying RL algorithms specifically for LLM agent training.

\newpage
\section{Working Examples}
\label{appendix_working_examples}
\begin{table}[h!]
\caption{Prompt template of Planner for text-based multi-hop question answering}
\centering
\resizebox{\textwidth}{!}{
\begin{tabular}{p{\textwidth}}
\toprule
\textbf{\#\# System Prompts \#\#} \\
**Your Task:** Answer the user's question thoroughly and accurately. \\

Before answering, outline your plan and reasoning process within `\textless think\textgreater ' tags. This includes understanding the question, identifying needed information, and planning your search queries. \\

You can use the search tool to get information. \\

Please put your search query in `\textless search\textgreater ' tags, e.g., `\textless search\textgreater Who is the mother of Xawery Żuławski? \textless /search\textgreater ' \\

Every `\textless search\textgreater ' tag only accepts one search query. \\

But you can use multiple `\textless search\textgreater ' tags if needed. \\

Note that your query should be explicit and specific. \\

It is difficult to search a query with reference. \\

For example, if you directly search `Who is the mother of the director of film Polish-Russian War', the search engine cannot find the information directly. \\

You should make a plan to decompose the question into multiple sub-queries. \\

Finally, integrate all the information and put your final answer within `\textless answer\textgreater ' tags. \\
\midrule
\textbf{\#\# User \#\#} \\
Question: Who succeeded the first President of Namibia? \\
\midrule
\textbf{\#\# Assistant \#\#} \\
\textless think\textgreater \\
The question asks about the person who succeeded the first President of Namibia. \\
1. I need to find who was the first President of Namibia. \\
2. I need to find who succeeded the first President of Namibia. \\
\textless /think\textgreater \\
\textless search\textgreater Who was the first President of Namibia? \textless /search\textgreater \\
\midrule
\textbf{\#\# User \#\#} \\
Obs: Sam Nujoma was the first President of Namibia. \\
\midrule
\textbf{\#\# Assistant \#\#} \\
\textless think\textgreater \\
Now I know that Sam Nujoma was the first President of Namibia. \\
I need to find who succeeded Sam Nujoma. \\
\textless /think\textgreater \\
\textless search\textgreater Who succeeded Sam Nujoma? \textless /search\textgreater \\
\midrule
\textbf{\#\# User \#\#} \\
Obs: Hifikepunye Pohamba succeeded Sam Nujoma as the President of Namibia. \\
\midrule
\textbf{\#\# Assistant \#\#} \\
\textless think\textgreater \\
The question asks about the person who succeeded the first President of Namibia. \\
Sam Nujoma was the first President of Namibia. \\
Hifikepunye Pohamba succeeded Sam Nujoma as the President of Namibia. \\
I have all the information to answer the question. \\
\textless /think\textgreater \\
\textless answer\textgreater 
Hifikepunye Pohamba 
\textless /answer\textgreater \\
\bottomrule
\end{tabular}
}
\label{planner_text}
\end{table}

\begin{table}[h!]
\caption{Prompt Template of Worker for Text-based Multi-hop Question Answering}
\centering
\resizebox{\textwidth}{!}{
\begin{tabular}{p{\textwidth}}
\toprule
\textbf{\#\# System Prompts \#\#} \\
You will be given a question with a context. \\
The context includes multiple passages labeled by index [x]. \\
Your task is to select the most relevant passage from the context based on the question. \\
Please state the reason why you choose the passage. Put your thinking process into the \textless think\textgreater ...\textless /think\textgreater tag. \\
Put the index of the selected passage into the \textless select\textgreater ...\textless /select\textgreater tag. \\
Based on the selected passage, you should answer the question. State the answer as a complete declarative sentence, mirroring the question structure. \\
Put your answer into the \textless sentence\textgreater ...\textless /sentence\textgreater tag. \\
If the passages have no information related to the question, put [-1] in the \textless select\textgreater ...\textless /select\textgreater tag \\
and try your best to answer the question according to your knowledge in the \textless sentence\textgreater ...\textless /sentence\textgreater tag. \\
\midrule
\textbf{\#\# User \#\#} \\
Question: What college did Kyeon Mi-ri attend?\\
Context:\\
$[0]$ Kyeon Mi-ri graduated from Seoul Traditional Arts High School in 1983, then studied Dance at Sejong University. She made her acting debut in 1984, and has since become active in television dramas, most notably as the arrogant and ambitious Lady Choi in the 2003 period drama "Dae Jang Geum" (or "Jewel in the Palace"), which was a hit not only in Korea but throughout Asia. \\
$[1]$ Shin Kyeong-nim was born on April 6, 1936 in North Chungcheong Province, South Korea. When he was young, Shin Kyong-rim frequented the people of Korea\'s rural villages and collected the traditional songs they sang. Much of his poetry represents a modernization of things he heard then Shin Kyeong-nim graduated in English Literature from Dongguk University, from which time he strove to become a creative writer. In 1955 and 1956, he made his formal literary debut with the publication of poems "Day Moon" (Natdal), "Reeds" (Galdae) and "Statue of Stone" (Seoksang). He taught elementary school in his hometown for a period of time, before moving to Seoul to work as an editor for Hyundae munhak and Donghwa Publishers. But his strong desire to create poetry continued. \\
$[2]$ Paige Ackerson-Kiely received a BA in Asian Studies from the University of New Mexico in Albuquerque. Prior to this achievement, she attended Beloit College in Beloit, Wisconsin, Marmara University in Istanbul, and Birzeit University in Birzeit, Palestine. \\
$[3]$ Han Seung-yeon was born on July 24, 1988, in Seoul, South Korea. She made her acting debut as a child actress in a bit part in "Dear Ends" (1993), "Summer Showers" (1995) and "Star in My Heart" (1997). She later left South Korea to study at Tenafly High School in New Jersey, United States. However, she withdrew from high school mid-course in order to pursue a singing career. After returning to South Korea, she debuted as a member of girl group Kara on March 29, 2007. During her time with the group, she passed a high school qualification exam, the College Scholastic Ability Test, and was accepted by Kyung Hee University, majoring in theater and film. \\
$[4]$ Myo Min Zaw studied English at the University of Yangon, where he became active in the pro-democracy group All Burma Federation of Student Unions (ABFSU). In December 1996, he participated in a student protest, and following the closing of Burma's universities, remaining involved in the pro-democracy movement. \\
\midrule
\textbf{\#\# Assistant \#\#} \\
\textless think\textgreater 
The question asks about what college Kyeon Mi-ri attended. \\
Passage [0] clearly states that "Kyeon Mi-ri graduated from Seoul Traditional Arts High School in 1983, then studied Dance at Sejong University." \\
So I select passage [0].
\textless /think\textgreater \\
\textless select\textgreater 
[0]
\textless /select\textgreater \\
\textless sentence\textgreater
Kyeon Mi-ri attended Sejong University.
\textless /sentence\textgreater \\
\bottomrule
\end{tabular}
}
\label{worker_text}
\end{table}

\begin{table}[h!]
\caption{Prompt Template of Planner for Knowledge-based Question Answering}
\centering
\resizebox{\textwidth}{!}{
\begin{tabular}{p{\textwidth}}
\toprule
\textbf{\#\# System Prompts \#\#} \\
Your task is answering questions. You will be given a question and some candidate entities. You should first make a plan about how to solve this question. \\
You can use the Search(entity, question) function to get information. \\
Please think thoroughly and label the final answer with \textless answer\textgreater *** \textless /answer\textgreater . \\
\midrule
\textbf{\#\# User \#\#} \\
Question: Which film came out first, Blind Shaft or The Mask Of Fu Manchu? \\
Candidate: [0] Blind Shaft [1] The Mask Of Fu Manchu\\
\midrule
\textbf{\#\# Assistant \#\#} \\
\textless think\textgreater \\
To solve this problem, I need to:\\
1. Figure out when Blind Shaft came out.\\
2. Figure out when The Mask Of Fu Manchu came out.\\
3. Compare their dates.\\
I need to search information for both of Blind Shaft and The Mask Of Fu Manchu.\\
\textless /think\textgreater \\
\textless action\textgreater \\
Search([0], "When did Blind Shaft come out?")\\
Search([1], "When did The Mask Of Fu Manchu come out?")\\
\textless /action\textgreater \\
\midrule
\textbf{\#\# User \#\#} \\
Obs: Blind Shaft came out on 2003 and The Mask Of Fu Manchu came out on 1932.\\
\midrule
\textbf{\#\# Assistant \#\#} \\
\textless think\textgreater Ok. Right now I need compare their released date. 1932 is much earlier than 2003. Therefore, The Mask Of Fu Manchu came out first. \textless /think\textgreater \\
\textless answer\textgreater The Mask Of Fu Manchu \textless /answer\textgreater \\
\bottomrule
\end{tabular}
}
\label{planner_kbqa}
\end{table}

\begin{table}[h!]
\caption{Prompt Template of Worker for Knowledge-based Question Answering}
\vspace{0.2cm}
\centering
\resizebox{\textwidth}{!}{
\begin{tabular}{p{\textwidth}}
\toprule
\textbf{\#\# System Prompts \#\#} \\
Task:
Extract relevant information from provided materials and present it in a specified format.\\
\\
Instructions:\\
\\
Read Carefully:\\
First, analyze the question to identify the key entity and requested information (e.g., "mother," "birth date").\\
Review all materials systematically.\\
\\
Extraction Rules:\\
Use only direct matches from materials. Never paraphrase or infer relationships.\\
If multiple materials contain the answer, select all of them.\\
If no material matches, respond with ``No relevant information found.''\\
\\
Answer Format:\\
State the answer as a complete declarative sentence, mirroring the question structure.\\
Add the source ID in brackets at the beginning of the answer.\\
\\
Use template:\\
\textless think\textgreater ... let's think step by step ... \textless /think\textgreater \\
\textless select\textgreater [X][Y] \textless /select\textgreater \\
\textless sentence\textgreater Full sentence answer \textless /sentence\textgreater \\
\midrule
\textbf{\#\# User \#\#} \\
Question: Who is the mother of Xawery Żuławski?\\
Materials:\\
$\text{[0]}$ Xawery Żuławski, mother, Małgorzata Braunek\\
$\text{[1]}$ Xawery Żuławski, father, Andrzej Żuławski\\
$\text{[2]}$ Xawery Żuławski, family, Q63532193\\
$\text{[3]}$ Xawery Żuławski, family name, Q56541485\\
$\text{[4]}$ Xawery Żuławski, spouse, Maria Strzelecka\\
$\text{[5]}$ Xawery Żuławski, date of birth, 1971-12-22T00:00:00Z\\
$\text{[6]}$ Xawery Żuławski, sibling, Vincent Zulawski\\
$\text{[7]}$ Xawery Żuławski, place of birth, Warsaw\\
$\text{[8]}$ Andrzej Żuławski, child, Xawery Żuławski\\
$\text{[9]}$ Małgorzata Braunek, child, Xawery Żuławski\\
\midrule
\textbf{\#\# Assistant \#\#} \\
\textless think\textgreater The question asks me to find the mother of Xawery Żuławski. [0] says Xawery Żuławski's mother Małgorzata Braunek, which excatly meets our need. \textless /think\textgreater\\
\textless select\textgreater [0] \textless /select\textgreater \\
\textless sentence\textgreater The mother of Xawery Żuławski is Małgorzata Braunek. \textless /sentence\textgreater \\
\bottomrule
\end{tabular}
}
\label{worker_kbqa}
\end{table}

\end{document}